# *The Socially Invisible Robot*
# Navigation in the Social World using Robot Entitativity

Aniket Bera[1], Tanmay Randhavane[1], Emily Kubin[2], Austin Wang[1], Kurt Gray[2], and Dinesh Manocha[1]

*Abstract*— We present a real-time, data-driven algorithm to enhance the *social-invisibility* of robots within crowds. Our approach is based on prior psychological research, which reveals that people notice and–importantly–react negatively to groups of social actors when they have high *entitativity*, moving in a tight group with similar appearances and trajectories. In order to evaluate that behavior, we performed a user study to develop navigational algorithms that minimize entitativity. This study establishes mapping between emotional reactions and multi-robot trajectories and appearances, and further generalizes the finding across various environmental conditions. We demonstrate the applicability of our entitativity modeling for trajectory computation for active surveillance and dynamic intervention in simulated robot-human interaction scenarios. Our approach empirically shows that various levels of entitative robots can be used to both avoid and influence pedestrians while not eliciting strong emotional reactions, giving multi-robot systems socially-invisibility.

## I. INTRODUCTION

As robots have become more common in social environments, people's expectations of their social skills have increased. People often want robots to be more socially visible–more salient social agents within group contexts [17]. This social visibility includes being more capable of drawing the attention of humans and evoking powerful emotions [22]. Cases of social visibility include tasks in which robots must work collaboratively with humans. However, not all contexts require socially visible robots. There are situations in which robots are not used to collaborate with people but instead used to monitor them. In these cases, it may be better for robots to be socially invisible.

*Social invisibility* refers to the ability of agents to escape the attention of other people. For example, psychological research reveals that African Americans often go unnoticed in social environments[11], especially reactions related to threat. Evolution has attuned the human brain to respond rapidly to threatening stimuli, thus the less a person–or a robot–induces negative emotion, the less likely it is to be noticed within a social milieu. The social invisibility conferred by not inducing emotion is especially important in surveillance contexts in which robots are expected to move seamlessly among people without being noticed. Noticing surveillance robots not only makes people hide their behavior, but the negative emotions that prompt detection may also induce reactance [9], which may lead to people to lash out and harm the robots or even other people [12] Research reveals a number of ways of decreasing negative emotional reactions towards social agents [10], but one element may be especially important for multi-robot systems: entitativity [13], "groupiness") is tied to three main elements, uniformity of appearance, common movement, and proximity to one another. The more agents look and move the same, and the closer agents are to each other, the more entitative a group seems, which is why a marching military platoon seems more grouplike than people milling around a shopping mall.

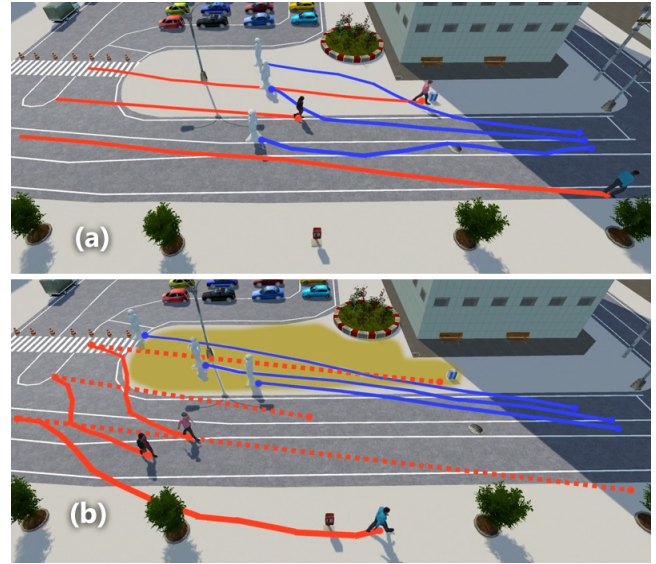

**Fig. 1:** *Multi-robot systems (robots marked by **blue** trajectories) are used among crowds for surveillance and monitoring. Our novel navigation algorithm takes into account various levels of physical and social constraints and use them for: (a) Active surveillance including monitoring crowds (**red** trajectories) while moving through them with no collisions; (b) Dynamic intervention where the robots try to influence the crowd behavior and movements and make the pedestrians avoid the area marked by a **yellow** overlay. The dashed **red** line indicates the predicted pedestrian trajectories if the robots did not attempt to dynamically intervene.*

The threatening nature of groups means that the more entitative (or grouplike) a collection of agents seem, the greater the emotional reaction they induce and the greater their social visibility. As maximizing the social invisibility of collections of agents requires minimizing perceptions of threat, it is important for multi-robot systems to minimize their entitativity. In other words, if multi-robots systems are to move through groups without eliciting negative reactions [16], they must seem more like individuals and less like a cohesive and coordinated group.

[1]Authors from the Department of Computer Science, University of North Carolina at Chapel Hill, USA
[2]Authors from the Department of Psychology and Neuroscience, University of North Carolina at Chapel Hill, USA

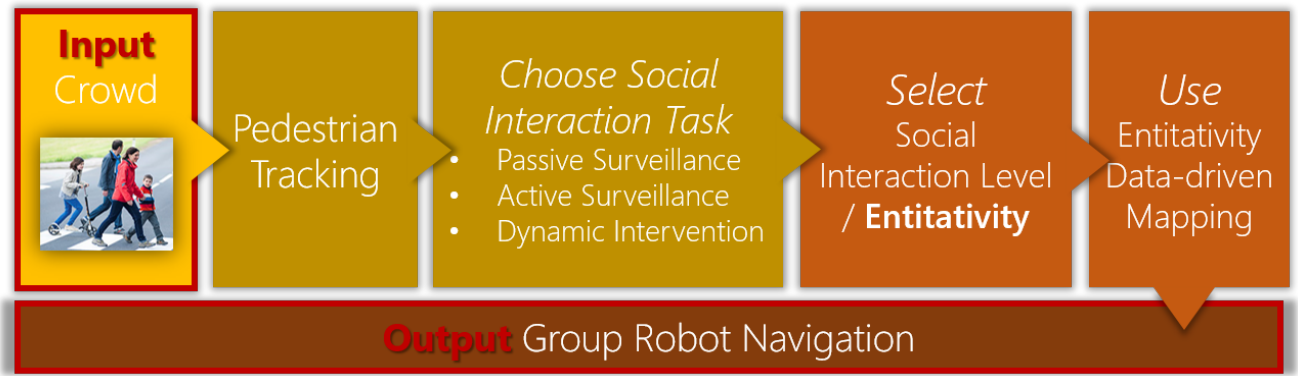

**Fig. 2:** *Our method takes a live or streaming crowd video as an input. We extract the initial set of pedestrian trajectories using an online pedestrian tracker. Based on the level of social invisibility we want to achieve, we compute motion model parameters of the robot group using a data-driven entitativity mapping (which we compute based on a user-study(Section IV)).*

**Main Results:** We present a novel, real-time planning algorithm that seeks to optimize entitativity within pedestrian environments in order to increase socially-invisible navigation (by minimizing negative emotional reactions). First, we conduct a user study to empirically tie trajectories of multi-robot systems to emotional reactions, revealing that–as predicted–more entitative robots are seen as more unnerving. Second, we generalize these results across a number of different environmental conditions (like lighting). Third, we extract the trajectory of each pedestrian from the video and use Bayesian learning algorithms to compute their motion model. Using entitativity features of groups of robots and the pedestrians, we perform long-term path prediction for the pedestrians. To determine these entitativity features we establish a data-driven entitativity mapping (EDM) between the group robot motion and entitativity measure from an elaborate web-based perception user study that compares the participants' emotional reactions towards simulated videos of multiple robots. Specifically, highly entitative collections of robots are reported as unnerving and uncomfortable. The results of our mapping are well supported by psychology literature on entitativity [34].

We highlight the benefits of our data-driven metric for use of multiple robots for crowd surveillance and active interference. We attempt to provide maximally efficient navigation and result in maximum social invisibility. In order to pursue different sets of scenarios and applications, we highlight the performance of our work in multiple surveillance scenarios based on the level of increasing social interaction between the robots and the humans.

Our approach has the following benefits:

**1. Entitativity Computation:** Our algorithm accurately predicts emotional reactions (entitativity) of pedestrians towards robots in groups.

**2. Robust computation:** Our algorithm is robust and can account for noise in pedestrian trajectories, extracted from videos.

**3. Fast and Accurate:** Our algorithm involves no pre-computation and evaluates the entitativity behaviors at interactive rates.

The rest of the paper is organized as follows. In Section 2, we review the related work in the field of psychology and behavior modeling. In Section 3, we give a background on quantifying entitativity and introduce our notation. In Section 4, we present our interactive algorithm, which computes the perceived group entitativity from trajectories extracted from video. In Section 5, we describe our user study on the perception of multiple simulated robots with varying degrees of entitativity.

## II. RELATED WORK

Human beings are inherently social creatures, making interacting with and perceiving others an important part of the human experience. Complex interactions within brain regions work harmoniously to navigate the social landscape [36]. Interesting patterns emerge when attempting to understand how humans view groups of people.

### A. Psychological Perspectives on Group Dynamics

A long-standing tenet of social psychology is that people's behaviors hinge upon their group context. Importantly, the impact of social dynamics is highly influenced by group contexts [38]–often for the worse. Decades of psychological research reveals that people interact more negatively with groups than with individuals [34], expressing more hostility towards and feeling more threatened by a group than an individual [16]. Such reactions to groups have real world implications, especially when onlookers have the ability to act violently. At the heart of these anti-social actions are negative emotional reactions, which can be directed at any social agent, whether human or robot [19]. Most often, these emotions are unease [8], threat [19], and fear [30].

### B. Human-Aware Robot Navigation

Many approaches have been applied towards the navigation of socially-aware robots [31], [7], [3], [15], [25], [29], [18],

[26], [24]. This type of navigation can be generated by predicting the movements of pedestrians and their interactions with robots [26], [4], [40], [33], [2]. Some algorithms use probabilistic models in which robots and human agents cooperate to avoid collisions [39]. Other techniques apply learning models which have proven useful in adapting paths to social conventions [27], [32], [35], [6]. Yet other methods model personal space in order to provide human-awareness [1]. This is one of many explicit models for social constraints [37], [23]. While these works are substantial, they do not consider psychological constraints or pedestrian personalities.

*C. Behavior Modeling of Pedestrians*

There is considerable literature in psychology, robotics, and autonomous driving on modeling the behavior of pedestrians [5], [15], [14]. Other techniques have been proposed to model heterogeneous crowd behaviors based on personality traits [20].

## III. SOCIAL INTERACTION

In this section, we present our interactive algorithm for performing socially-invisible robot navigation in crowds. Our approach can be combined with almost any real-time pedestrian tracker that works on dense crowd videos. Figure 2 gives an overview of our approach. Our method takes a live or streaming crowd video as an input. We extract the initial set of pedestrian trajectories using an online pedestrian tracker. Based on the level of social invisibility we want to achieve, we find motion model parameters of the robot group using a data-driven entitativity mapping (which we compute based on a user-study(Section IV)).

*A. Entitativity*

Entitativity is the perception of a group comprised of individuals as a single entity. People sort others into entities like they group together objects in the world, specifically by assessing common fate, similarity, and proximity [13]. When individuals are connected by these properties, we are more likely to perceive them as a single entity. Larger groups are more likely to be perceived as entities, but only when there is similarity among the groups individual members [28].

Entitativity is the extent to which a group resembles a single entity versus of collection of individuals; in other words, it is the groups "groupiness" or "tightness" [13], [21]. Overall, entitativity is driven by the perception of three main elements:

**1. Uniformity of appearance:** Highly entitative groups have members that look the same.

**2. Common movement:** Highly entitative groups have members that move similarly.

**3. Proximity:** Highly entitative groups have members that are very close to each other.

*B. Notation and Terminology*

Notation used in the rest of the paper will be presented in this section. A collection of agents is called a crowd. The agents in a crowd are called *pedestrians*, while the agents that must navigate through a crowd are called *robots*. Each agent has a *state* describing its trajectory and movement parameters. These parameters dictate the agent's movement on a 2D plane. An agent's state is defined as $\mathbf{x} \in \mathbb{R}^6$:

$$\mathbf{x} = [\mathbf{p} \ \mathbf{v}^c \ \mathbf{v}^{pref}]^\mathbf{T}, \quad (1)$$

where $\mathbf{p}$ is the agent's position, $\mathbf{v}^c$ is its current velocity, and $\mathbf{v}^{pref}$ is the *preferred velocity* on a 2D plane. The preferred velocity describes the velocity that the agent takes if there are no other agents or obstacles nearby. In real-world situations, other agents and obstacles in an agent's vicinity cause a difference between $\mathbf{v}^{pref}$ and $\mathbf{v}^c$, which indicates the degree of the agent's interactions with its environment. The current state of the environment, denoted by $S$, describes the states of all other agents and the current positions of obstacles in the scene. The state of the crowd, defined as the union of each pedestrian's state, is represented as $\mathbf{X} = \bigcup_i \mathbf{x_i}$, where subscript $i$ denotes the $i^{th}$ pedestrian. Within a crowd, pedestrians can be partitioned into groups (also called clusters) based on their motion. We represent a group of pedestrians as $\mathbf{G} = \bigcup_j \mathbf{x_j}$ where subscript $j$ denotes the $j^{th}$ pedestrian in the group.

The motion model is the local navigation rule or scheme that each agent uses to avoid collisions with other agents or obstacles and has a group strategy. The parameters of the motion model is denoted $\mathbf{P} \in \mathbb{R}^6$. We based our model on the RVO velocity-based motion model [41]. In this model, the motion of each agent is governed by these five individual pedestrian characteristics: *Neighbor Dist, Maximum Neighbors, Planning Horizon, (Radius) Personal Space, and Preferred Speed* and one group characteristic: *Group Cohesion*. We combine RVO with a group navigation scheme in Section 4.2. In our approach, we mainly analyze four parameters ($\mathbf{GP} \in \mathbb{R}^4$): *Neighbor Dist, (Radius) Personal Space, Group Cohesion, and Preferred Speed*.

**Entitativity Metric:** Prior research in psychology takes into account properties such as uniformity, common movement, and proximity, and models the perception of *entitativity* using the following 4-D feature vector:

$$\mathbf{E} = \begin{pmatrix} Friendliness \\ Creepiness \\ Comfort \\ Unnerving \end{pmatrix} \quad (2)$$

*Friendliness, Creepiness, Comfort* and *Unnerving (ability to unnerve)* are the emotional impressions made by the group on observers. Using Cronbach's $\alpha$ (a test of statistical reliability) in pilot studies we observed that the parameters were highly related with $\alpha = 0.794$, suggesting that they were justifiable adjectives for socially-invisible navigation.

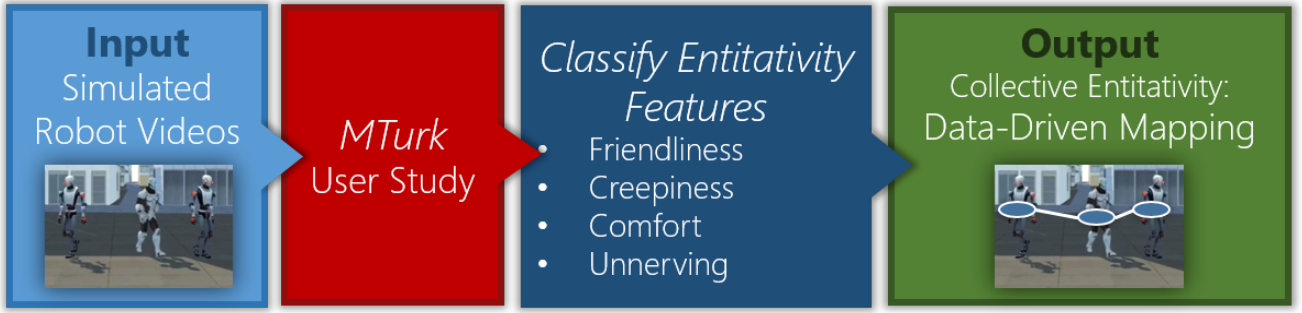

**Fig. 3:** *We generate a variety of simulated robot groups using our motion model. Based on our precomputed EDM, we classify the entitativity of each cluster or group of pedestrians based on the four components: friendliness, creepiness, unnerving, and comfort.*

## IV. Data-Driven Robot Entitativity (EDM)

In order to evaluate the impact of the group motion model parameters on the perception of entitativity of a group of robots, we performed a user study using simulated trajectories. We provide the details of this user study in this section.

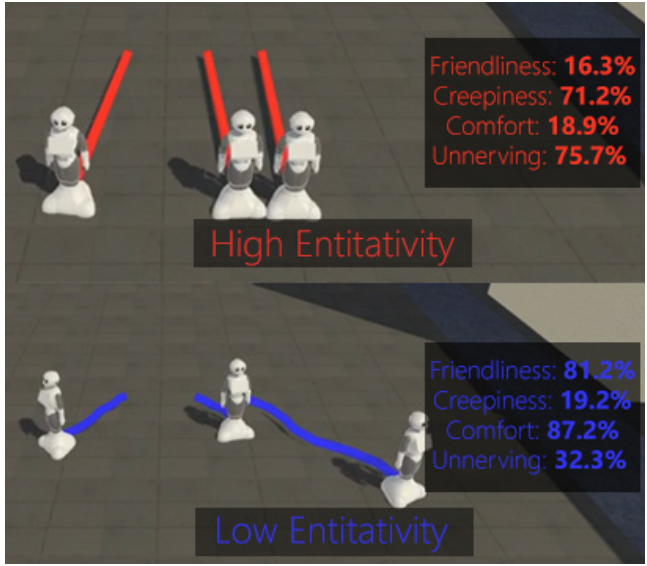

**Fig. 4:** *Varying Levels of Entitativity: Parameters of the group motion model affected the entitativity of multiple simulated robots. Agents having the same speed and similar trajectories were perceived to be highly entitative (top) whereas agents walking at different speeds and varying trajectories were perceived as less entitative (bottom).*

### A. Study Goals

The aim of this study was to understand how the perception of multiple pedestrians is affected by the parameters of the group motion model. We use the results of this user study to compute a statistical EDM between the group motion model parameters and the perception of groups in terms of friendliness, creepiness, and social comfort.

### B. Experimental Design

Here, we provide details of the design of our experiment.

*1) Participants:* We recruited 100 participants (78 male, $\bar{x}_{age} = 32.85$, $s_{age} = 10.10$) from Amazon Mechanical Turk.

*2) Procedure:* A web-based study was performed in which participants were asked to watch pairs of simulated videos of robots and compare their entitativity features. Each video contained 3 simulated agents with various settings of the group motion model parameters. We consider variations in four group motion models parameters(**GP**): *Neighbor Dist, Radius, Pref Speed, and Group Cohesion*. In each pair, one of the videos corresponds to the default values of the parameters. The other video was generated by varying one parameter to either the minimum or the maximum value. Thus each participant watched 8 pairs of videos corresponding to the minimum and the maximum value for each motion model parameter. The participants watched the two videos side by side in randomized order and compared the entitativity features of the robots groups in the two videos. They could watch the videos multiple times if they wished. Demographic information about participants' gender and age was collected at the beginning of the study.

| *Parameters* (**GP**) | *min* | *max* | *default* |
|---|---|---|---|
| Neighbor Distance ($m$) | 3 | 10 | 5 |
| Radius (Personal Space) ($m$) | 0.3 | 2.0 | 0.7 |
| Preferred speed ($m/s$) | 1.2 | 2.2 | 1.5 |
| Group Cohesion | 0.1 | 1.0 | 0.5 |

**TABLE I:** *Default values for simulation parameters used in our experiments*

*3) Questions:* For each trial, the participant compared the two videos (*Left* and *Right*) on a 5-point scale from Left (-2) - Right (2). We used the following questions to record participants' responses on friendliness, creepiness, or social comfort experienced:

1) In which of the videos did the characters seem more friendly?
2) In which of the videos did the characters seem more creepy?
3) In which of the videos did you feel more comfortable around the characters?
4) In which of the videos did you feel more unnerved by the characters' movement?

These questions were motivated by previous studies [34]. We define an entitativity feature corresponding to each question.

Thus we represent the entitativity features of a group as a 4-D vector: *Friendliness, Creepiness, Unnerving (Ability to Unnerve), Comfort*.

*C. Analysis*

We average the participant responses to the each video pair to obtain 8 entitativity feature data points ($\mathbf{E}_i, i = 1, 2, ..., 8\}$). Table II provides the correlation coefficients between the questions for all the participant responses. The high correlation between the questions indicates that the questions measure disjoint aspects of a single perception feature, entitativity. As expected, *creepiness* and *unnerving* are inversely correlated with *friendliness* and *comfort*. Principal Component Analysis of the four entitativity features also reveals that only 2 principal components are enough to explain over 96% of the variance in the participants' responses. We still use the four features instead of the principal components because they provide more interpretability.

|              | *Friendliness* | *Creepiness* | *Comfort* | *Unnerving* |
|--------------|----------------|--------------|-----------|-------------|
| *Friendliness* | 1            | -0.829       | 0.942     | -0.802      |
| *Creepiness*   | -0.829       | 1            | -0.906    | 0.858       |
| *Comfort*      | 0.942        | -0.906       | 1         | -0.833      |
| *Unnerving*    | -0.802       | 0.858        | -0.833    | 1           |

**TABLE II:** ***Correlation Between Questions:*** *We provide the correlation coefficients between the questions. The high correlation between the questions indicates that the questions measure different aspects of a single perception feature, entitativity.*

We vary the motion model parameters one by one between their high and low values (while keeping the other parameters at default value). Given the entitativity features obtained using the psychology study for each variation of the motion model parameters, we can fit a generalized linear model to the entitativity features and the model parameters. For each video pair $i$ in the gait dataset, we have a vector of parameter values and a vector of entitativity features $\mathbf{E}_i$. Given these parameters and features, we compute the entitativity mapping of the form:

$$\begin{pmatrix} Friendliness \\ Creepiness \\ Comfort \\ Unnerving \end{pmatrix} = \mathbf{G_{mat}} * \begin{pmatrix} \frac{1}{14}(Neighbor\ Dist - 5) \\ \frac{1}{3.4}(Radius - 0.7) \\ \frac{1}{2}(Pref.\ Speed - 1.5) \\ \frac{1}{1.8}(Group\ Cohesion - 0.5) \end{pmatrix} \quad (3)$$

We fit the matrix $\mathbf{G_{mat}}$ using generalized linear regression with each of the entitativity features as the responses and the parameter values as the predictors using the normal distribution:

$$\mathbf{G_{mat}} = \begin{pmatrix} -1.7862 & -1.0614 & -2.1983 & -1.7122 \\ 1.1224 & 1.1441 & 1.7672 & -0.2634 \\ -1.0500 & -1.2176 & -2.1466 & -0.9220 \\ 1.1948 & 1.7000 & 0.9224 & 0.3622 \end{pmatrix}. \quad (4)$$

We can make many inferences from the values of $\mathbf{G_{mat}}$. The negative values in the first and third rows indicate that as the values of motion model parameters increase, the friendliness of the group decreases. That is, fast approaching and cohesive groups appear to be less friendly. This validates the psychological findings in previous literature [34]. One interesting thing to note is that creepiness increases when group cohesion decreases. When robots walk in a less cohesive group, they appear more creepy but they may appear less unnerving.

We can use our data-driven entitativity model to predict perceived entitativity of any group for any new input video. Given the motion parameter values $\mathbf{GP}$ for the group, the perceived entitativity $\mathbf{E}$ can be obtained as:

$$\mathbf{E} = \mathbf{G_{mat}} * \mathbf{GP} \quad (5)$$

*D. Socially-Invisible Navigation*

To provide socially-invisible navigation, we use the entitativity level of robots. We control the entitativity level depending on the requirements of the social-invisibility. We represent the social-invisibility as a scalar $s \in [0, 1]$ with $s = 0$ representing very low social-invisibility and $s = 1$ representing highly socially-invisible robots. Depending on the applications and situations, the social-invisibility can be varied.

We relate the desired social-invisibility ($s$) to entitativity features $\mathbf{E}$ as follows:

$$s = 1 - \frac{\|\mathbf{E} - \mathbf{E}_{min}\|}{\|\mathbf{E}_{max} - \mathbf{E}_{min}\|} \quad (6)$$

where $\mathbf{E}_{max}$ and $\mathbf{E}_{min}$ are the maximum and minimum entitativity values obtained from the psychology study.

According to Equation 6, there are multiple entitativity features $\mathbf{E}$ for the desired social-invisibility $s$. This provides flexibility to choose which features of entitativity we wish to adjust and we can set the desired entitativity $\mathbf{E}_{des}$ that provides the desired social-invisibility level. Since $\mathbf{G_{mat}}$ is invertible (Equation 5), we can compute the motion model parameters $\mathbf{GP}_{des}$ that achieve the desired entitativity:

$$\mathbf{GP}_{des} = \mathbf{G_{mat}}^{-1} * \mathbf{E}_{des} \quad (7)$$

These motion model parameters $\mathbf{GP}_{des}$ are the key to enabling socially-invisible collision-free robot navigation through a crowd of pedestrians. Our navigation method is based on Generalized Velocity Obstacles (GVO) [42], which uses a combination of local and global methods. The global metric is based on a roadmap of the environment. The local method computes a new velocity for each robot and takes these distances into account. Moreover, we also take into account the dynamic constraints of the robot in this formulation - for example, mechanical constraints that prevent the robot from rotating on the spot.

At a given time instant, consider a robot $i$ with position $\mathbf{p}^c_{robot_i}$ and preferred velocity $\mathbf{v}^{pref}_{robot_i}$ (Figure 6). The preferred velocity is computed from the global navigation module of GVO and represents the velocity it would have for navigating to its goal position in the absence of social constraints. In each time step, it must choose a velocity that navigates it closer to its (current) goal while remaining as socially invisible as permissible. If it were to use the predicted positions $\mathbf{p}^{pred}_{human}$ of pedestrians to update its own velocity to $\mathbf{v}^{pred}_{robot_i}$, it would certainly avoid collisions with both pedestrians and scene obstacles, but may fail at its assigned task. For example, if a robot is tasked with preventing pedestrians from encroaching on a demarcated zone, it is not enough to predict their positions in the upcoming time step and update its own velocity accordingly. We therefore sacrifice some level of social invisibility by increasing the entitativity of the robots so as to dynamically intervene in pedestrian movement. The aim in such a scenario is to induce pedestrians to walk away from the restricted zone by presenting them with a more entitative group of robots. Concretely, we use the motion model parameters $\mathbf{GP}_{des}$ discussed earlier to compute a goal position $\mathbf{p}_{robot_i}^{pred+inv}$ for the pedestrian and a new velocity $\mathbf{v}_{robot_i}^{pred+inv}$ for the robot. The robot velocity $\mathbf{v}_{robot_i}^{pred}$ computed from the nave approach may lead to pedestrians intruding on restricted zones, whereas the velocity $\mathbf{v}_{robot_i}^{pred+inv}$ computed from our entitative approach will prevent this while crucially maintaining a desired level of social invisibility for the robots.

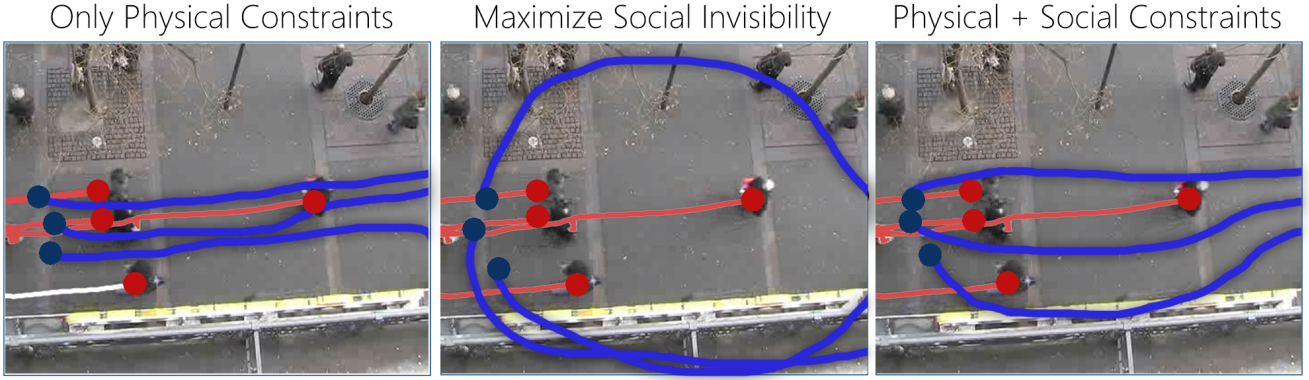

**Fig. 5:** *Example robot trajectory* navigating through the crowd in the **Hotel** dataset. **Red** circles/trajectories represent current pedestrian positions and **blue** circles/trajectories are the current positions of the robot. The left figure computes paths based on only physical constraints (collision-handling, smooth trajectory computation, sensor errors and uncertainty, and dynamics) at the cost of higher social constraints (zero social invisibility); the middle figure minimizes social constraints (i.e. entitativity) at the cost of longer paths; the right figure (our algorithm) balances social and physical constraints, and computes appropriate trajectories for the robot (in **blue**).

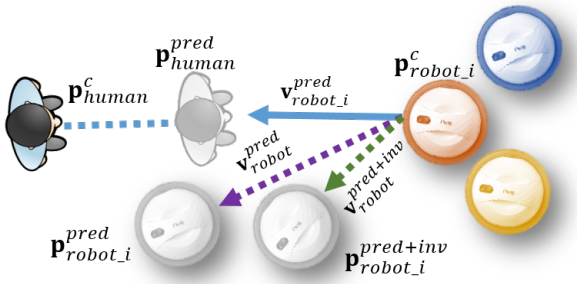

**Fig. 6:** *To provide socially-invisible navigation, we use the entitativity level of robots. We control the entitativity level depending on the requirements of the social-invisibility.*

## V. APPLICATIONS

We present some driving applications of our work that are based on use of multi-robot systems for crowd surveillance and control. In these scenarios, our method optimizes multi-robot systems so that they can interact with such crowds seamlessly based on physical constraints (e.g. collision avoidance, robot dynamics) and social invisibility. We simulate our algorithm with two sets of surveillance scenarios based on the level of increasing social interaction between the robots and the humans:

*1) Active Surveillance:* This form of patrolling or surveillance includes autonomous robots that share a physical space with pedestrians. While performing surveillance and analysis, these robots will need to plan and navigate in a collision-free manner in real-time amongst crowds. In this case, the robots need to predict the behavior and trajectory of each pedestrian. For example, marathon races tend to have large populations, with a crowd whose location is constantly changing. In these scenarios, it is necessary to have a surveillance system that can detect shifting focal points and adjust accordingly.

In such scenarios, the robots need to be highly socially-invisible ($s = 0$). We achieve this by setting the entitativity features to the minimum $\mathbf{E} = \mathbf{E}_{min}$ (Equation 6).

*2) Dynamic intervention:* In certain scenarios, robots will not only share a physical space with people but also influence pedestrians to change or follow a certain path or behavior. Such interventions can either be overt, such as forcing people to change their paths using visual cues or pushing, or subtle (for example, nudging). This type of surveillance can be used in any scenario with highly dense crowds, such as a festival or marathon. High crowd density in these events can lead to stampedes, which can be very deadly. In such a scenario, a robot can detect when density has reached dangerous levels and intervene, or "nudge" individuals until they are distributed more safely.

For dynamic intervention with pedestrians or robits, we manually vary the entitativity level depending on urgency or agent proximity to the restricted area. In these situations, we restrict the entitativity space by imposing a lower bound

$s_{min}$ on the social-invisibility (Equation 6):

$$s_{min} \leq 1 - \frac{\|\mathbf{E} - \mathbf{E}_{min}\|}{\|\mathbf{E}_{max} - \mathbf{E}_{min}\|}. \tag{8}$$

*A. Performance Evaluation*

We evaluate the performance of our socially-invisible navigation algorithm with GVO [42], which by itself does not take into account any social constraints. We compute the number of times a pedestrian intrudes on a designated restricted space, and thereby results in issues related to navigating through a group of pedestrians. We also measure the additional time that a robot with our algorithm takes to reach its goal position, without the pedestrians intruding a predesignated restricted area. Our results (Table III) demonstrate that in $< 30\%$ additional time, robots using our navigation algorithm can reach their goals while ensuring that the restricted space is not intruded. Table III also lists the time taken to compute new trajectories while maintaining social invisibility. We have implemented our system on a Windows 10 desktop PC with Intel Xeon E5-1620 v3 with 16 GB of memory.

| *Dataset* | *Additional Time* | *Intrusions Avoided* | *Performance* |
|---|---|---|---|
| NPLC-1 | 14% | 3 | 3.00E-04 ms |
| NDLS-2 | 13% | 2 | 2.74E-04 ms |
| IITF-1 | 11% | 3 | 0.72E-04 ms |
| NDLS-2 | 17% | 4 | 0.98E-04 ms |

**TABLE III:** *Navigation Performance for Dynamic Intervention: A robot using our navigation algorithm can reach its goal position, while ensuring that any pedestrian does not intrude the restricted space with $< 15\%$ overhead. We evaluated this performance in a simulated environment, though the pedestrian trajectories were extracted from the original video. In all the videos we have manually annotated a specific area as the restricted space.*

We have also applied our algorithm to perform active surveillance (Table IV). The pedestrian density in these crowd videos varies from low-density (less than 1 robot per square meter) to medium-density (1-2 robots per square meter), to high-density (more than 2 robots per square meter).

| *Dataset* | *Analyzed Pedestrians* | *Input Frames* | *Performance* |
|---|---|---|---|
| IITF-1 | 15 | 450 | 2.70E-04 ms |
| IITF-3 | 27 | 238 | 7.90E-04 ms |
| IITF-5 | 25 | 450 | 8.30E-04 ms |
| NPLC-1 | 17 | 238 | 3.80E-04 ms |
| NPLC-3 | 42 | 450 | 1.80E-04 ms |
| NDLS-2 | 38 | 238 | 1.90E-04 ms |
| Manko | 16 | 373 | 1.01E-03 ms |
| Marathon | 27 | 450 | 9.10E-04 ms |
| Explosion | 28 | 238 | 5.80E-04 ms |
| Street | 67 | 9014 | 1.0E-05 ms |

**TABLE IV:** *Navigation Performance for Active Surveillance: Performance of our entitativity computation on different crowd videos for performing active surveillance. We highlight the number of video frames used for extracted trajectories, and the running time (in milliseconds).*

## VI. CONCLUSIONS, LIMITATIONS AND FUTURE WORK

Drawing from work in social psychology, we develop a novel algorithm to minimize entitativity and thus maximize the social invisibility of multi-robot systems within pedestrian crowds. A user-study confirms that different entitativity profiles–as given by appearance, trajectory and spatial distance–are tied to different emotional reactions, with high entitativity groups evoking negative emotions in participants. We then use trajectory information from low-entitative groups to develop a real-time navigation algorithm that should enhance social invisibility for multi-robot systems.

Our approach has some limitations. Although we did generalize across a number of environmental contexts, we note that motion-based entitativity is not the only feature involved in social salience and other judgments. People use a rich set of cues when forming impressions and emotionally reacting to social agents, including perceptions of race, class, religion, and gender. As our algorithm only uses motion trajectories, it does not exhaustively capture all relevant social features. However, motion trajectories are an important low-level feature of entitativity and one that applies especially to robots, who may lack these higher-level social characteristics.

Future research should extend this algorithm to model the appearances of robots in multi-robot systems. Although many social cues may not be relevant to robots (e.g., race), the appearance of robots can be manipulated. Research suggests that robots that march will have higher entitativity and hence more social visibility. This may prove a challenge to manufacturers of surveillance robots, as mass production typically leads to identical appearances. Another key future direction involves examining the interaction of the perceiver's personality with the characteristics of multi-robot systems, as some people may be less likely to react negatively to entitative groups of robots, perhaps because they are less sensitive to general threat cues or, more specifically, have more experience with robots.